\documentclass[10pt]{article}

\usepackage{arxiv}

\usepackage[utf8]{inputenc} 
\usepackage[T1]{fontenc}    
\usepackage{hyperref}       
\usepackage{url}            
\usepackage{booktabs}       
\usepackage{amsfonts}       
\usepackage{nicefrac}       
\usepackage{microtype}      
\usepackage{lipsum}		
\usepackage{graphicx}
\usepackage{natbib}
\usepackage{doi}

\usepackage{soul}
\usepackage{times}
\usepackage{epsfig}
\usepackage{graphicx}
\usepackage{amsmath}
\usepackage{amssymb}
\usepackage{url}
\usepackage{amsmath,graphicx}
\usepackage{float}
\usepackage{svg}
\usepackage{booktabs}
\usepackage{url}
\usepackage{subfig}
\usepackage{rotating}
\newcommand\tab[1][1cm]{\hspace*{#1}}
\usepackage{enumitem}

\usepackage{url}
\usepackage{lipsum}  
\usepackage{textcomp}
\usepackage{rotating}
\usepackage{ dsfont }
\usepackage{amsfonts}
\usepackage{textcomp}
\usepackage{gensymb}
\usepackage{adjustbox}
\usepackage{graphicx}
\usepackage{graphics}
\usepackage{algorithm}
\usepackage{booktabs}
\usepackage{filecontents}
\usepackage{tikz,pgfplots}
\usepackage{nameref}
\usepackage{varioref}
\usepackage{hyperref}
\usepackage{cleveref}
\usepackage{algpseudocode}
\newcommand{\comment}[1]{}
\usepackage{adjustbox}
\usepackage{graphicx}
\usepackage{array,booktabs,arydshln,xcolor}
\newcommand\VRule[1][\arrayrulewidth]{\vrule width #1}
\setcitestyle{numbers}

\title{Synthetic Latent Fingerprint Generator}

\author{{Andre Brasil Vieira Wyzykowski} \\
	Department of Computer Science and Engineering\\
	Michigan State University\\
	\And
	{Anil K. Jain} \\
	Department of Computer Science and Engineering\\
	Michigan State University\\
}

\renewcommand{\shorttitle}{}

\hypersetup{
pdftitle={Synthetic Latent Fingerprint Generator},
pdfsubject={q-bio.NC, q-bio.QM},
pdfauthor={Andre Brasil Vieira Wyzykowski, Anil Kumar Jain},
pdfkeywords={SYNTHETIC LATENT FINGERPRINT, Latents},
}

\begin{document}
\maketitle

\begin{abstract}
Given a full fingerprint image (rolled or slap), we present CycleGAN models to generate multiple latent impressions of the same identity as the full print. Our models can control the degree of distortion, noise, blurriness and occlusion in the generated latent print images to obtain Good, Bad and Ugly latent image categories as introduced in the NIST SD27 latent database. The contributions of our work are twofold: (i) demonstrate the similarity of synthetically generated latent fingerprint images to crime scene latents in NIST SD27 and MSP databases as evaluated by the NIST NFIQ 2 quality measure and ROC curves obtained by a SOTA fingerprint matcher, and (ii) use of synthetic latents to augment small-size latent training databases in the public domain to improve the performance of DeepPrint, a SOTA fingerprint matcher designed for rolled to rolled fingerprint matching on three latent databases (NIST SD27, NIST SD302, and IIITD-SLF). As an example, with synthetic latent data augmentation, the Rank-1 retrieval performance of DeepPrint is improved from 15.50\% to 29.07\% on challenging NIST SD27 latent database. Our approach for generating synthetic latent fingerprints can be used to improve the recognition performance of any latent matcher and its individual components (e.g., enhancement, segmentation and feature extraction).
\end{abstract}

\section{Introduction}

Since the first use of fingerprints in criminal investigations in 1891, fingerprints have become the most widely used tool for accurately and quickly identifying perpetrators \cite{caplan1990fingerprints}. The frequent use of fingerprints in judicial courts worldwide has demonstrated their acceptance as scientific evidence in the conviction of criminals. Within this context, latent fingerprints - those collected from various surfaces in crime scene investigations, are essential for identifying suspects involved in a crime. The identification of these latent prints by comparing them against rolled/slap fingerprints in law enforcement databases continues to be crucial for criminal investigations.

\begin{figure}[H]
\centering
    \setlength{\tabcolsep}{3pt}
            \begin{tabular}{ccc}

            \includegraphics[height=4.5cm]{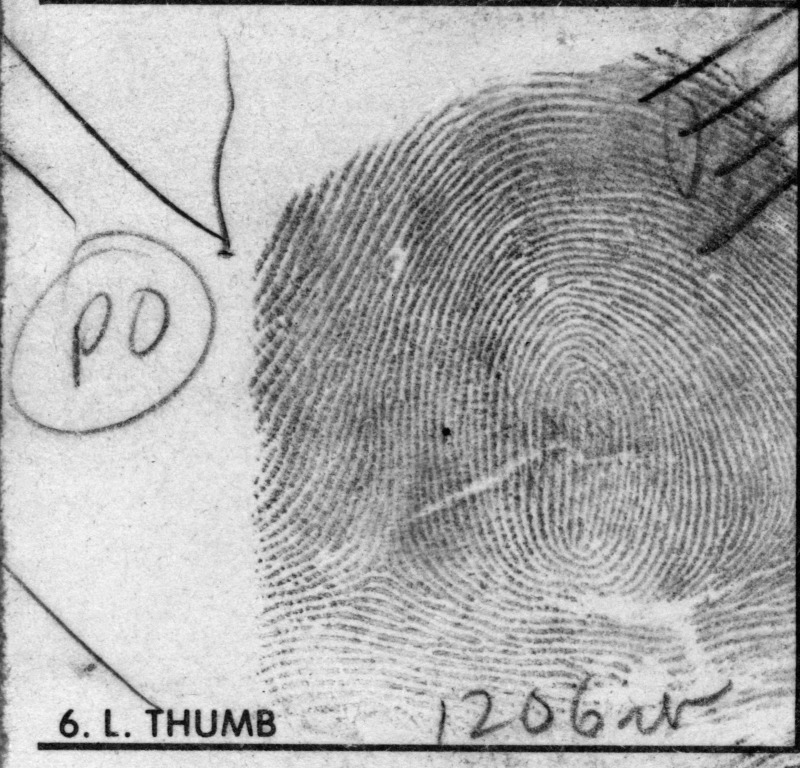}&
            \includegraphics[height=4.5cm]{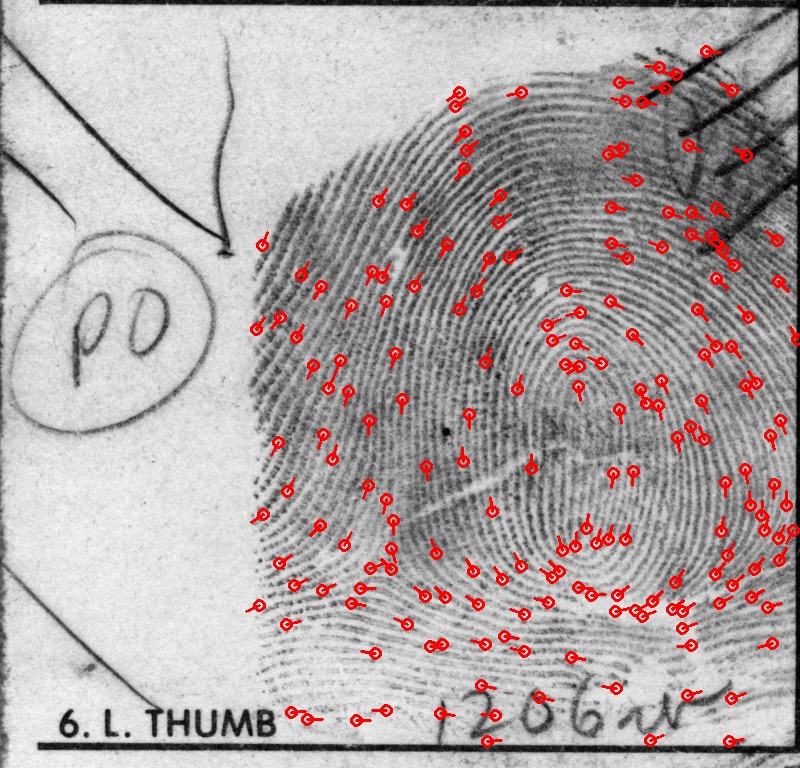}&
            \includegraphics[height=4.5cm]{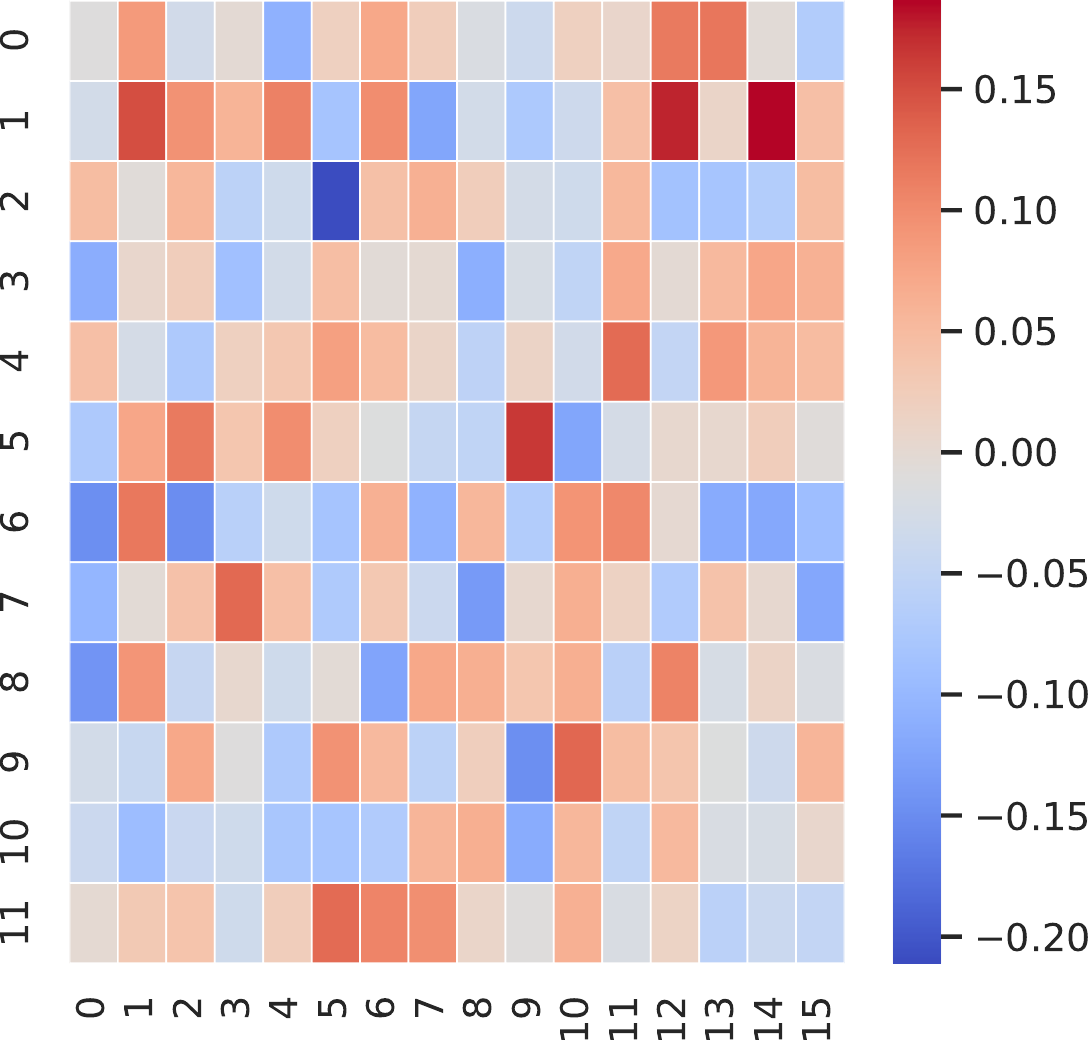}\\
            
            \includegraphics[height=4.5cm]{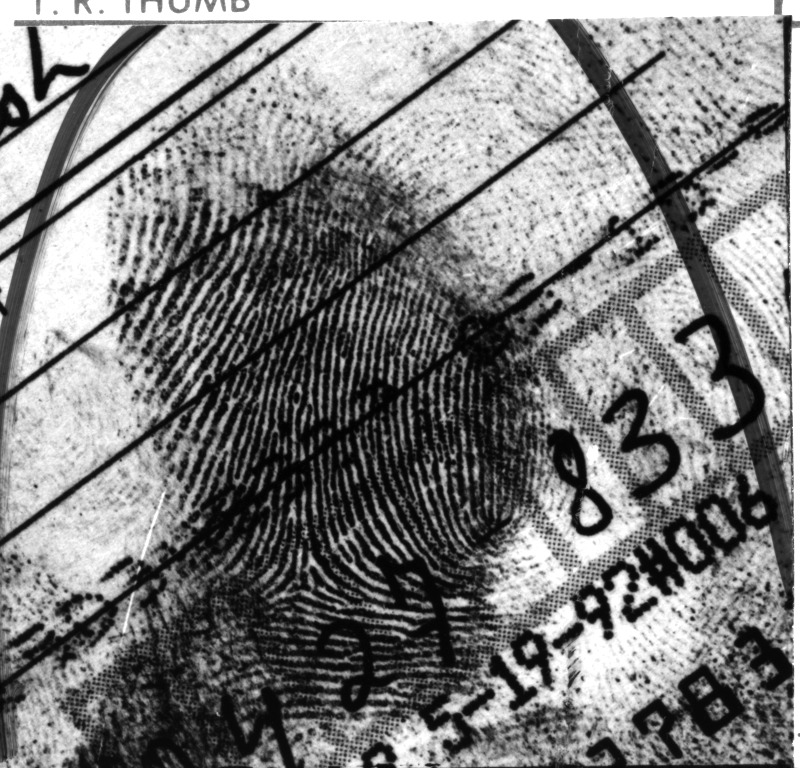}&
            \includegraphics[height=4.5cm]{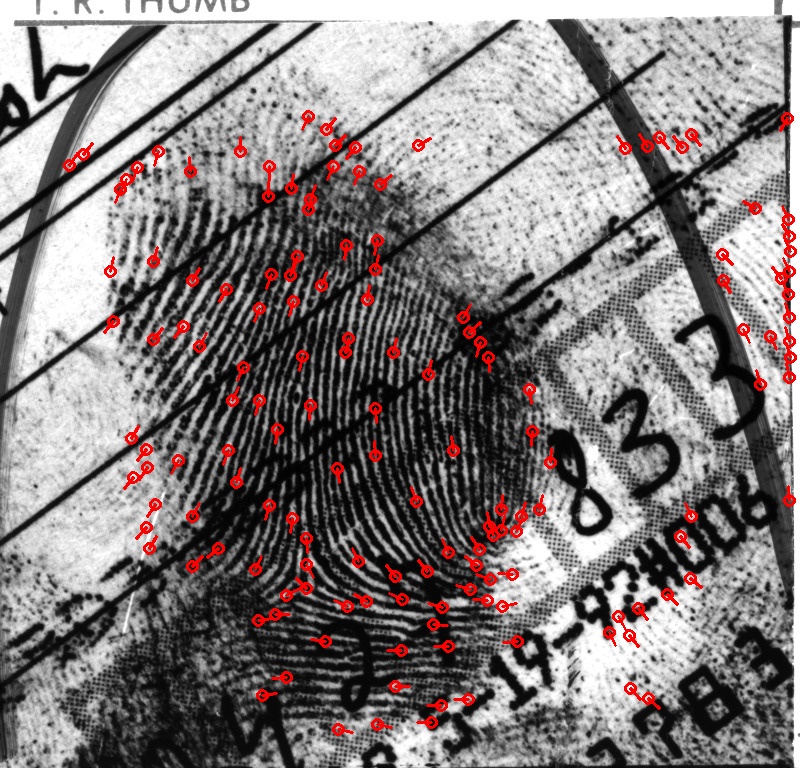}&
            \includegraphics[height=4.5cm]{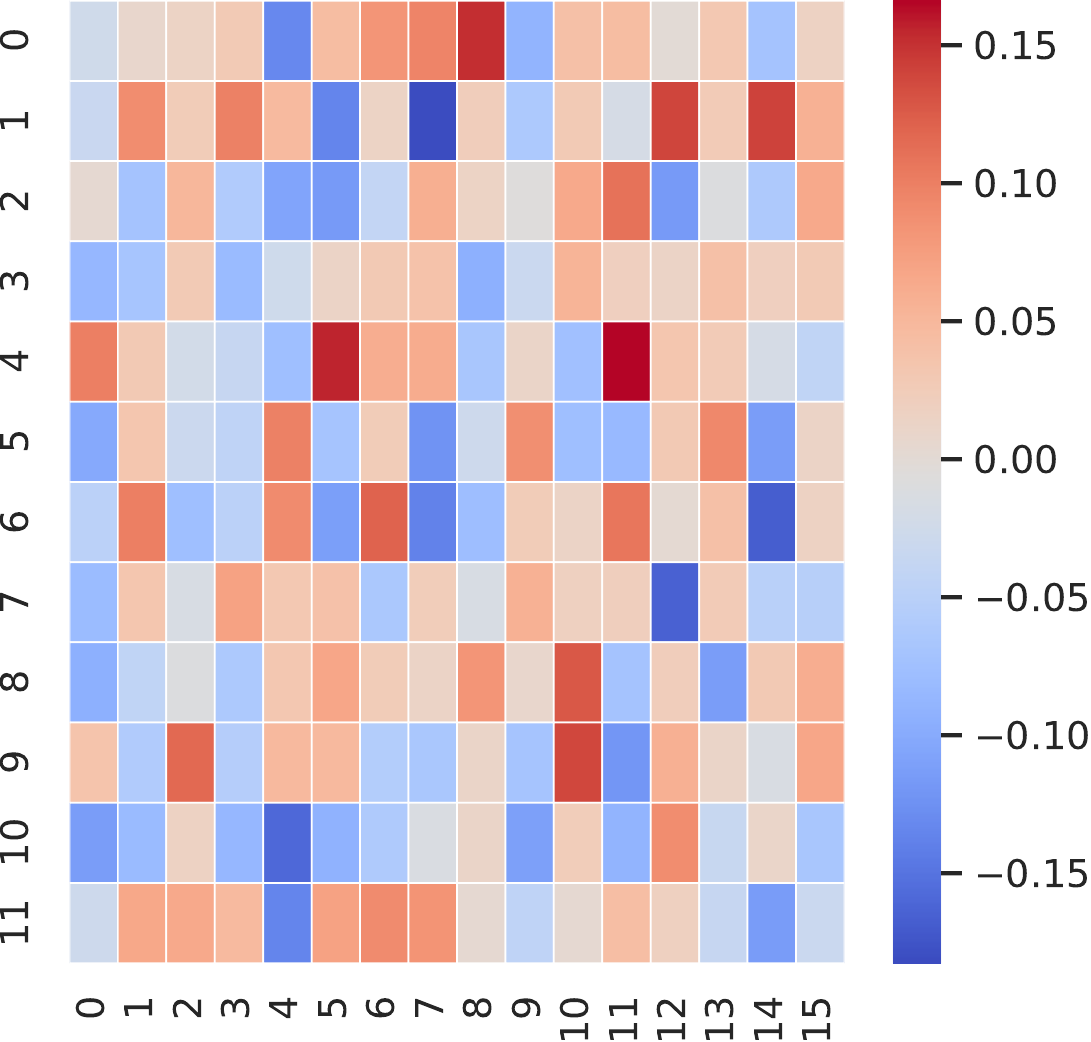}\\

            \end{tabular}
            \caption{Examples of a latent fingerprint (bottom row) and its rolled fingerprint mate (top row) from the NIST special database 27 (SD27) \cite{garris2000nist}. The second column shows the fingerprints with their minutiae points representation, and the last column shows the deep-network embeddings (192-dimensional) by DeepPrint \cite{engelsma2019learning} represented as a heatmap. }
            \label{sd27examples}
\end{figure}

Latent fingerprint images often contain various sources of noise, including blood, contaminants, and natural secretions. Furthermore, fingers leave distinct traces on surfaces, such as cellulose composites, glass, metal, and plastic, which lead to variations in the latent images captured in forensic data collection\cite{yamashita2011latent}. In addition, each surface where a person leaves their fingerprints provides distinct background for friction ridge impressions, such as printed or handwritten text, as illustrated in Figure \ref{sd27examples}.

Given the poor quality of latent fingerprints, the most challenging problem in fingerprint recognition is to perform identification using latent prints. Given a query latent print, it needs to be compared with large collections of rolled and plain fingerprints in forensic databases. Jain and Feng \cite{jain2010latent} designed a latent-to-rolled
matching algorithm based on minutiae, singular points (delta and core), and ridge flow map. Later, Cao et al. \cite{cao2019end} improved the recognition accuracy by implementing a deep-learning-based representation to enhance poor-quality latent images prior to minutiae extraction. These studies demonstrate that there are many open challenges, primarily related to matching poor-quality latent prints with relatively high-quality rolled/slap fingerprints captured under supervision during booking. Indeed, NIST ELFT evaluation \cite{flanagan2010nist} showed that the best performing latent matching algorithm was able to provide only 67\% Rank-1 accuracy in retrieving the mate of a query latent with a gallery of 100K rolled prints. When we compare this with rolled-to-rolled matching accuracy of over 99\%, then it is clear that latent fingerprint recognition offers fertile grounds for research \cite{wilson2004fingerprint}. 

\begin{figure*}[!h]
	\centering
 \setlength{\tabcolsep}{2pt}
		\begin{tabular}{cc|cc|cc|cc}
		
            \includegraphics[height=2cm]{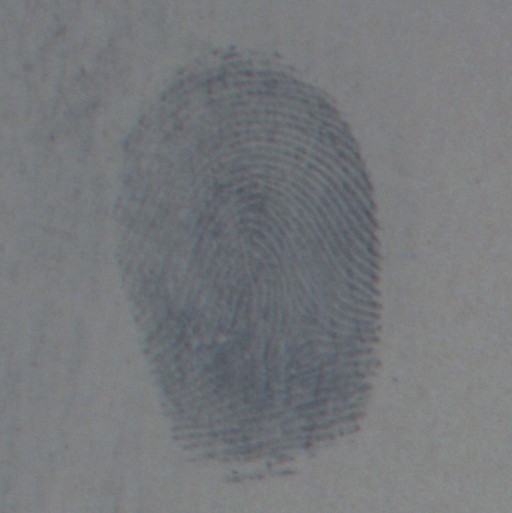} & 
            \includegraphics[height=2cm]{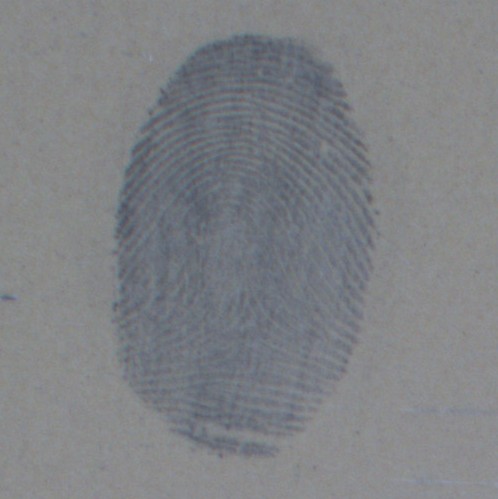} &  

            \includegraphics[height=2cm]{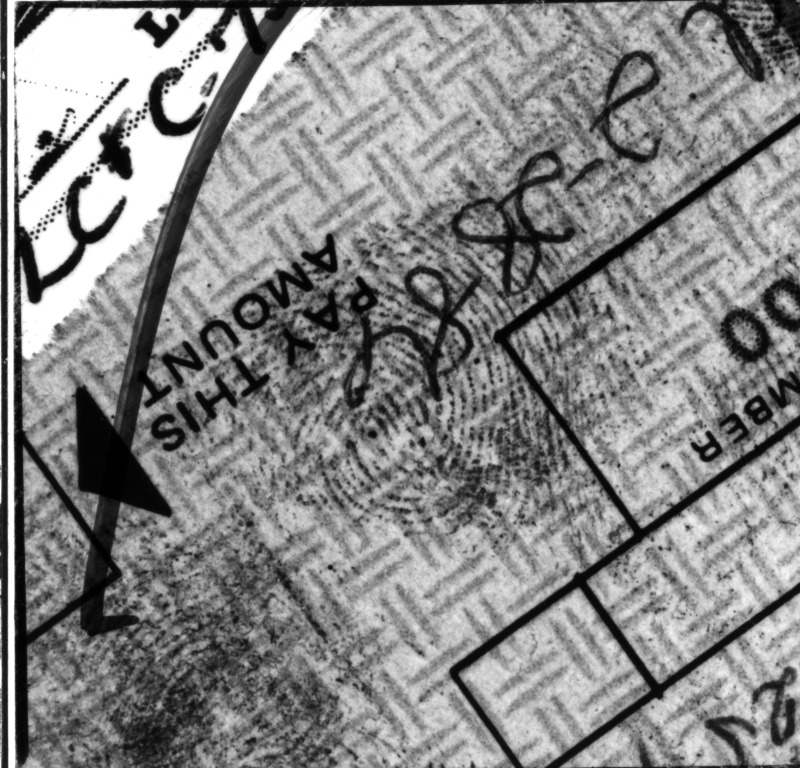} &  
            \includegraphics[height=2cm]{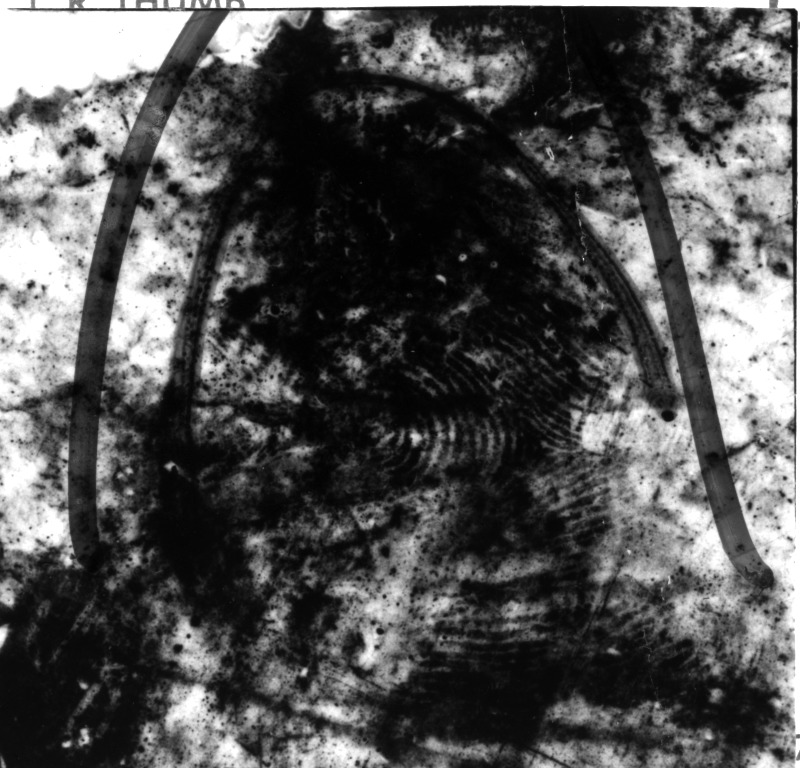} &  

            \includegraphics[height=2cm]{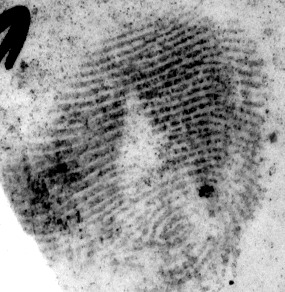} &  
            \includegraphics[height=2cm]{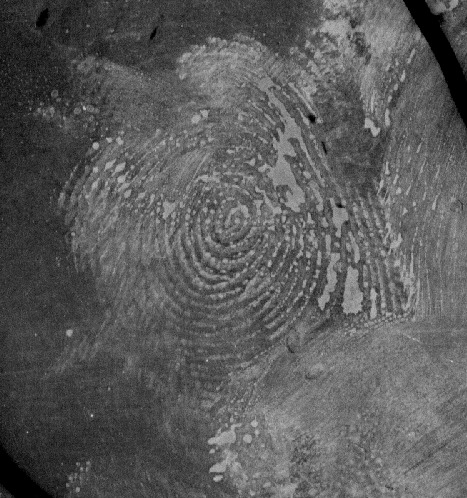} &  

             \includegraphics[height=2cm]{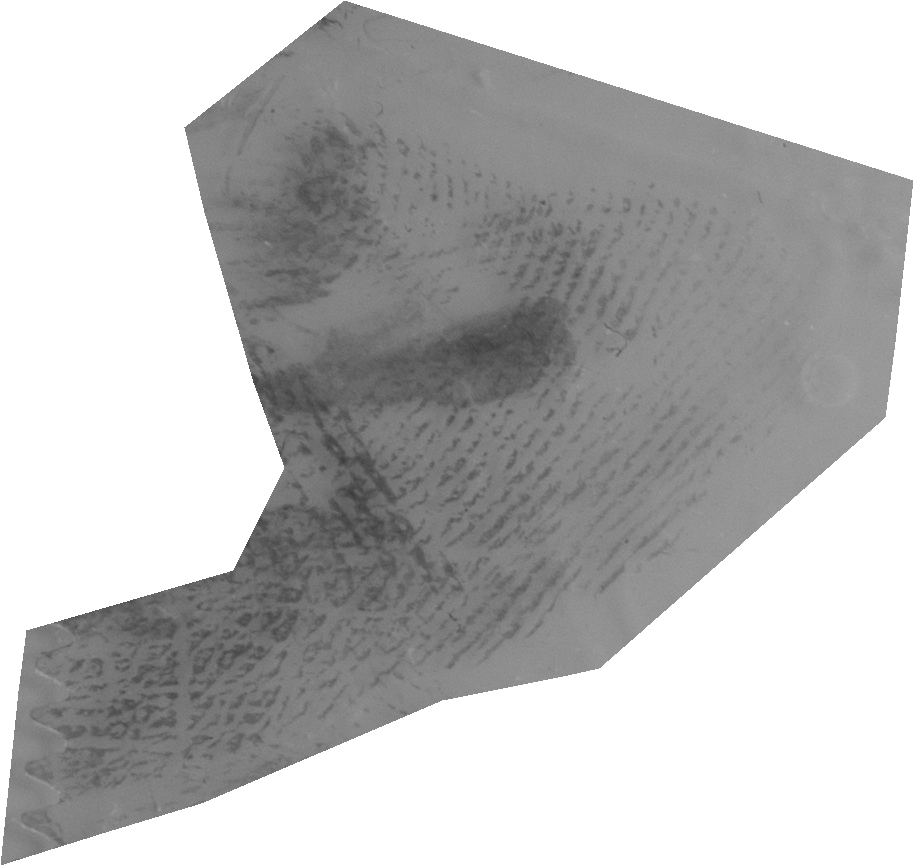} &  
            \includegraphics[height=2cm]{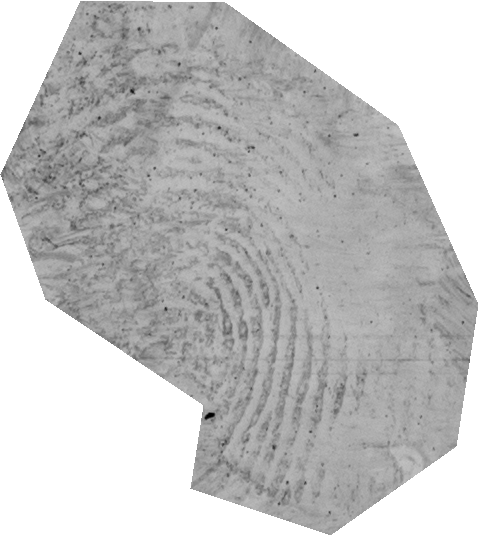} 

            \end{tabular}

                     \tab[0.58cm] IIITD-SLF \cite{yusof2012multi} \tab[1.8cm] NIST SD27 \cite{garris2000nist} \tab[1.6cm]  MSP Latent DB \cite{yoon2015longitudinal}  \tab[1.5cm]  NIST SD302 \cite{fiumara2019nist}

          \caption{Latent fingerprint examples from IIITD-SLF, NIST SD27, MSP latent and NIST SD302.}
    \label{latent_finger_examples} 
\end{figure*}

Within this context, several factors make it difficult to conduct research on latent fingerprint recognition: 

\begin{enumerate}
    \item Sparsity of public domain databases available for training and evaluating a latent fingerprint recognition system. Table \ref{tableDatabases} contains a summary of the latent fingerprint databases which have been used in academic publications. Figure \ref{latent_finger_examples} shows latent fingerprint examples from some of these databases. Note that the only two latent databases that are in the public domain are NIST SD302 and IIITD-SLF which were collected in a laboratory setting and not from crime scenes. One of the most popular latent fingerprint database in academic publications is NIST SD27 \cite{garris2000nist}, but it was withdrawn from the public domain by NIST. While NIST SD302 and IIITD-SLF are useful for research, they are relatively small and do not capture the type of distortion and background that is observed in operational databases such as NIST SD27 (see Figures \ref{sd27examples} and \ref{latent_finger_examples}).
    \item Limited variations in the style and quality of the latent images in public domain databases. These factors include small friction ridge area, small no. of minutiae points, blurred regions, and background complexity. The study by Gonzalez et al. \cite{loyola2021impact} demonstrates the challenges posed by such variations in latent fingerprint recognition accuracy; detection of even a small no. of spurious minutiae or failure to detect a few genuine minutiae can drastically impact the latent recognition accuracy. Given the complexity of the latent to rolled fingerprint matching problem, the lack of publicly available operational latent fingerprint databases has hindered the progress in latent fingerprint recognition despite its critical role in law enforcement and forensics \cite{national2009strengthening, thompson2017forensic}. According to the Innocence Project\footnote{\url{https://innocenceproject.org/forensic-science-problems-and-solutions/}}, ``Forensic science, or more specifically, problems in forensic science, contributes to many wrongful convictions, as seen in nearly half (45\%) of DNA exoneration cases and one-quarter (24\%) of all exonerations in the United States." and this includes errors in latent fingerprints matching.

\end{enumerate}

\begin{table}[H]
\centering
\caption{Latent Fingerprint databases used in this study.}
\label{tableDatabases}
\resizebox{0.7\columnwidth}{!}{%

\begin{tabular}{@{}lccc@{}}
\toprule
\multicolumn{1}{c}{\textbf{Database}}                                            & \textbf{\begin{tabular}[c]{@{}c@{}}Public \\ domain\end{tabular}} & \textbf{\begin{tabular}[c]{@{}c@{}}\# of unique fingers\\ (\# rolled-latent pairs)\end{tabular}} & \textbf{Collection details}                                                                                                 \\ \midrule
NIST SD27\cite{garris2000nist}                                                                        & No                                                                       & 258                                                                          & \begin{tabular}[c]{@{}c@{}} Crime scene images\end{tabular}                                  \\
MSP Latent DB \cite{yoon2015longitudinal}                                                                               & No                                                                       & 1,910                                                                          & \begin{tabular}[c]{@{}c@{}}Crime scene images \end{tabular}                                  \\
NIST SD302 \cite{fiumara2019nist}                                                                       & Yes                                                                      & 1,019*                                                                       & \begin{tabular}[c]{@{}c@{}}Laboratory collection\end{tabular}                 \\

IIITD-SLF \cite{yusof2012multi}                                                                        & Yes                                                                      & 150                                                                          & \begin{tabular}[c]{@{}c@{}}Laboratory collection\end{tabular} \\ \bottomrule
\end{tabular}
}\\
\scriptsize *Obtained after filtering the rolled and latent mates present in the finger position annotation in the SD302h subset. The total no. of latent fingerprints is 9,990. 
\end{table}

The objective of this research is to develop a method to generate a large collection of realistic synthetic latent fingerprint images that would be publicly available to interested researchers. The purpose of latent fingerprint synthesis is to advance pre-processing of latent images, including: (i) enhancement and segmentation,  (ii) data augmentation for learning a robust representation, and (iii) designing a robust matcher for latent to rolled matching.

With these objectives, this paper aims to design and develop models to generate a large collection of synthetic latent fingerprints from a given set of rolled prints. For each rolled print, we are able to generate multiple latent prints with different quality levels (Good, Bad and Ugly following the categorization of NIST SD27 database). We validate our synthesis approach by fine-tuning a state-of-the-art fingerprint matcher,  DeepPrint \cite{engelsma2019learning} with synthesized images. We show that this data augmentation leads to improved performance of DeepPrint on latent fingerprint recognition on NIST SD27, NIST SD302 and IIITD-SLF latent databases.

Our contributions in this paper are as follows:
\begin{enumerate}[noitemsep]
    \item  A method to generate synthetic latent fingerprints of different difficulties in terms of recognition.
    \item Demonstrate, both qualitatively and quantitatively, the similarity of our synthetic latent fingerprints to real latents.
    \item Utility of synthetic latents for data augmentation to improve the performance of a state-of-the-art deep network model for latent fingerprint recognition.
\end{enumerate}

Note that our objective here is not to build a state-of-the-art latent fingerprint recognition system but to present models for synthetic latents generation to assist in designing a SOTA latent matcher. For this reason, we do not address issues related to pre-processing of latents, including enhancement and segmentation, typically done in COTS latent matchers.

\section{Background}

We first describe general fingerprint synthesis approaches in Section \ref{sythBack}, followed by methods for synthesizing latent fingerprints in Section \ref{latentandReconstruction}.

\subsection{Fingerprint synthesis}
\label{sythBack}

One of the earliest and most prominent fingerprint synthesis generator is SFinGe\cite{article_sfinge}, which used handcrafted methods to generate ridge patterns, minutiae, and textures formed by the ridge-valley pattern. But due to the limitations of the model used as well as the computational resources at the time, the images lacked visual realism. 

With the introduction of generative adversarial networks (GAN) \cite{goodfellow2014generative}, the next generation of synthetic fingerprint approaches \cite{cao2018fingerprint, mistry2020fingerprint, bahmani2021high} emerged which are capable of generating plain and rolled fingerprints. However, like SFinGe, they are not designed for latent fingerprint synthesis. Also, these generators only create a single impression per finger. 

Engelsma et al. \cite{engelsma2022printsgan} proposed a state-of-the-art approach called PrintsGAN capable of generating a large number of distinct identities with multiple rolled/plain impressions per finger. Although PrintsGAN has shown promise for data augmentation, some minutiae are lost, or spurious ones are introduced during the creation of multiple impressions of an identity. Like PrintsGAN, Grosz and Jain \cite{grosz2022spoofgan} developed a model capable of synthesizing multiple fingerprint spoof impressions from the same identity. 

Other fingerprint synthesis approaches use CycleGAN \cite{zhu2017unpaired} to generate realistic textures. Wyzykowski et al. \cite{wyzykowski2021level, wyzykowski2022multiresolution} used CycleGAN to generate additional medium and high-resolution 500 ppi and 1250 ppi fingerprints images, corresponding to the SFinGe-based identity generator. Sams et. al \cite{sams2022hq} consolidate the use of CycleGAN, enabling the generation of new identities with StyleGAN 2 \cite{karras2020analyzing}. Despite these advances, all of the aforementioned approaches did not focus on latent fingerprint synthesis. 

\subsection{Latent fingerprint synthesis and reconstruction}
\label{latentandReconstruction}

Ozturk et al. \cite{ozturk2022minnet} developed an algorithm for automated latent fingerprint recognition. Although the paper's focus is not on latent fingerprint synthesis, the authors used a private dataset of latent fingerprints to generate a model with CycleGAN capable of converting images from FVC databases of plain fingerprints \cite{maio2004fvc2004} into latent prints. However, the authors did not use synthetic latent data for data augmentation purposes to show the utility of their synthesized latent database. Further, they did not (i) provide the number of identities generated, and (ii) evaluate the similarity between synthetic and real fingerprint latents.

Xu et al. \cite{xu2020augmentation} demonstrated that they can improve matching performance by using data augmentation in latent image reconstruction. However, the authors did not specify the metrics for evaluating latent fingerprints reconstructions, such as a minutiae detection analysis and their image quality.

\section{Proposed Latent Fingerprint Generator}

\begin{figure*}[h]
   \centering
     \includegraphics[height=4.8cm]{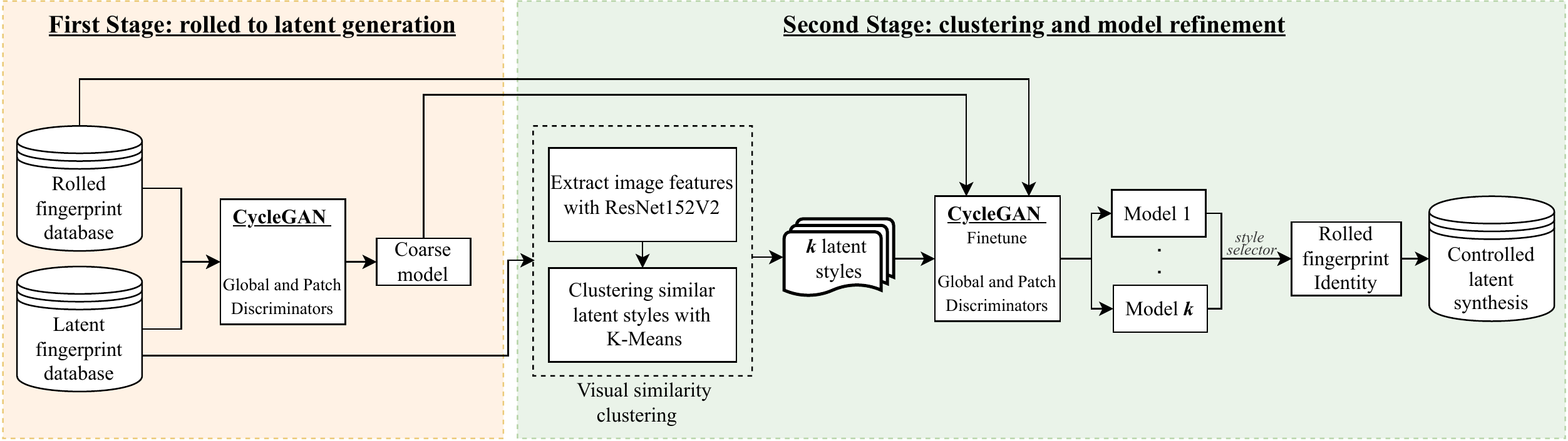}
    \caption{Steps to create synthetic latent fingerprints using the proposed approach.}
    \label{Flowchart} 
\end{figure*}

This study's main objective is to create latent fingerprints having control of the style generated by these models. This feature is essential since latent databases vary depending on the material or surface where the prints were left. We detail our method in Figure \ref{Flowchart} with the pseudocode in Algorithm \ref{alg:cap} below.

\begin{algorithm}
\caption{Latent fingerprint synthesis}\label{alg:cap}

\begin{algorithmic}
\renewcommand{\algorithmicrequire}{\textbf{Input:}}
\Require  $M(i)$ set of rolled prints.
\renewcommand{\algorithmicrequire}{\textbf{Output:}}
\Require Synthetic latent prints from $M(i)$ denoted as $SLP (i)$.


\renewcommand{\algorithmicrequire}{\textbf{First stage:}}

\Require 

\State Generate a coarse CycleGAN style model $GSM$ using mated rolled and latent prints as training data.

\State \textit{output: $GSM$ CycleGAN model}

\renewcommand{\algorithmicrequire}{\textbf{Second stage:}}

\Require \textit{input: ($GSM$ CycleGAN model)}

\State Use ResNet152V2 \cite{he2016identity} to extract features from the latent images.
\State Use K-Means for clustering the latent prints.
\For{each $k$ cluster}
    \State Generate a fine-tuned CycleGAN style model $SM_k$ using rolled and clustered latent prints as training data.
\EndFor
\For{each CycleGAN $SM_k$ model}
    \For{each rolled prints in $M(i)$}
        \State Generate $SLP$ synthetic latents prints.
\EndFor

    \EndFor
\State \textit{output: $SLP (i)$ synthetic latents prints.}

\end{algorithmic}

\end{algorithm}

In Section \ref{CycleGANDetails} we detail the modifications and parameters used in CycleGAN in both stages of our method. In Section \ref{section1}, we describe the first stage of the fingerprint synthesis method, where we create a generic model capable of transforming an image from the rolled domain into a generic latent domain. Then, Section \ref{section2} details the second stage, where we create specific models for each latent print style. 

\subsection{CycleGAN modification details}
\label{CycleGANDetails}

\begin{figure}[H]
   \centering
     \includegraphics[width=12.3cm]{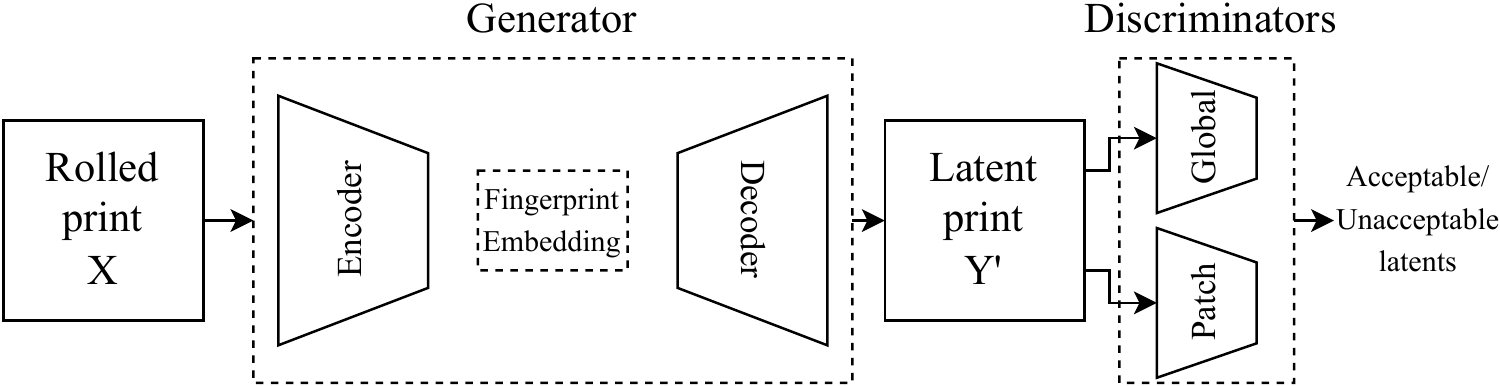}
    \caption{CycleGAN architecture with discriminators.}
    \label{CycleGAN} 
\end{figure}

CycleGAN \cite{ZhuPIE17} is a neural network model focused on the process of unpaired image-to-image translation, i.e., a network able to map two distinct image domains and perform the transformation of image features from these domains. Through the Cycle-Consistency loss, CycleGAN can perform two-way style translation, transforming input images into styles of output images and vice versa. Furthermore, CycleGAN does not require paired examples to train, an important advantage when working with latent images, since there is no pairing between the latent and rolled impression in many databases. However, we noticed stability problems during training when we performed our initial training with the original CycleGAN architecture on latent prints. Other authors \cite{hu9improved, kwon2021cycle} also report similar problems using CycleGAN on different image styles. Thus, we decided to use a global discriminator (already present in CyleGAN) and a patch discriminator, which we detail in Figure \ref{CycleGAN}. The patch discriminator stabilized our training, generating better details of the fingerprint ridges. We use a public implementation of CycleGAN\footnote{\url{https://github.com/towardsautonomy/CycleGAN_improved}} to perform the modifications described in this section.

We used Leaky Relu \cite{DBLP:journals/corr/XuWCL15} as the activation function in the generator with a negative slope of $\alpha = 0.2$. For the global and patch discriminators, we keep the original CycleGAN activation function, Relu. Additionally, we also used instance normalization.

There are six residual blocks \cite{he2016deep} in our CycleGAN architecture. We train our model using the Adam optimizer \cite{kingma2014adam} with $\beta_1 = 0.5$ and $\beta_2 = 0.999$. Our  weight for cycle-consistency loss was $\lambda = 10$. We used a learning rate of 0.0003 for the generator and 0.0001 for the discriminator. We early-stop \cite{prechelt1998early} the training if the loss does not improve after 50 epochs. 

Our work aims to transform rolled fingerprints into latent fingerprints, not to reverse the process (rolled into latent). However, by utilizing Cycle-Consistent Loss, CycleGAN can improve latent images, which may further enhance the performance of fingerprint recognition systems, an investigation we plan to undertake in future work. 

As described in Sections \ref{section1} and \ref{section2}, we use our modified version of CycleGAN for stages 1 and 2.

\subsection{First stage: rolled to latent generation}
\label{section1}

This first stage aims to create a CycleGAN ``coarse" model capable of mapping latent and rolled domains. Using this ``coarse" model as input, improved fine-tuned models are created in the second stage (see Section \ref{section2}).

\subsection{Second stage: clustering and model refinement}
\label{section2}

A latent fingerprint image can have different styles depending on where the fingerprint traces were left or if noise, dirt, blood, etc., were present. Therefore, reproducing this inherent variation is vital. Thus, the second stage involves clustering latent prints and generating custom CycleGAN models tailored to the styles of each cluster. First, in Section \ref{cluster}, we detail our approach to clustering latent fingerprints. Then, Section \ref{finetuneCycleGAN} describes the finetuning process in CycleGAN, generating models capable of creating styles corresponding to clusters. 

\subsubsection{Cluster latent images based on visual similarity}
\label{cluster}

We first used a pre-trained public ResNet152V2 \cite{he2016identity}\footnote{\url{https://www.tensorflow.org/api_docs/python/tf/keras/applications/resnet_v2/ResNet152V2}} model to extract feature vectors of the latent images. This neural network was trained using the ImageNet-21K dataset \cite{ridnik2021imagenet}, providing good classification performance. We use the last fully-connected layer of ResNet152V2 with 2048 output nodes as a feature vector to cluster the images with K-Means. As a result of this process, we obtain $k$ sets of latent images, each containing distinct patterns and visual features. These $k$ clusters then serve as input to perform a finetuning with CycleGAN in stage 2 (see Section \ref{finetuneCycleGAN}).

\begin{figure*}[!ht]
	\centering
		\begin{tabular}{cc!{\VRule[1pt]}cc!{\VRule[1pt]}cc}
			\begin{turn}{90}\footnotesize \tab[0.55cm] NIST SD27\end{turn}
            \includegraphics[height=2.25cm]{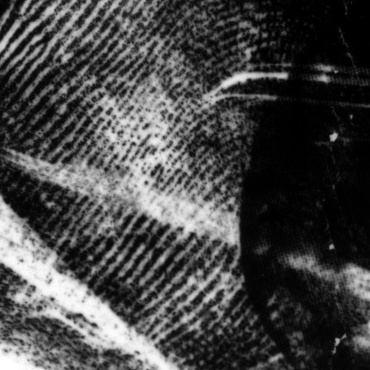} &  
             \includegraphics[height=2.25cm]{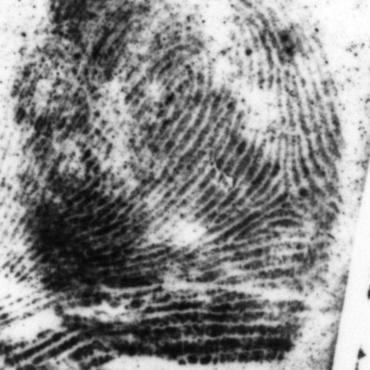}  &
             \includegraphics[height=2.25cm]{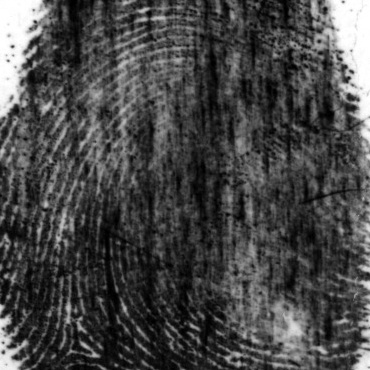} &
            \includegraphics[height=2.25cm]{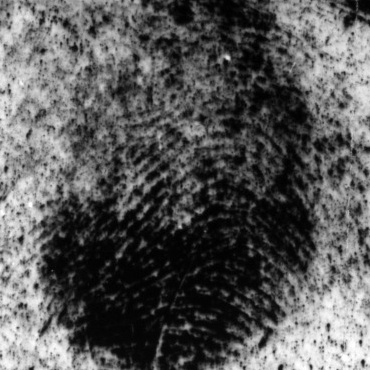}&
            \includegraphics[height=2.25cm]{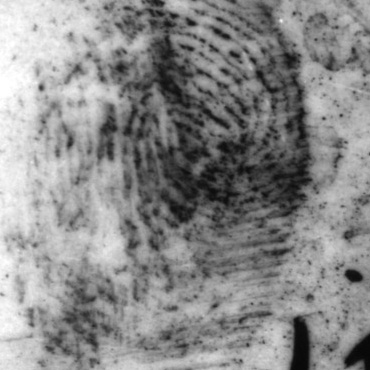} &
            \includegraphics[height=2.25cm]{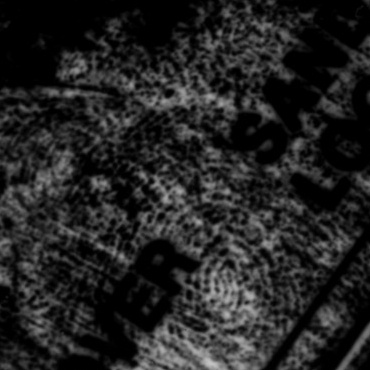}
             \\
             
            \begin{turn}{90}\footnotesize \tab[0.55cm] MSP latent\end{turn}
            \includegraphics[height=2.25cm]{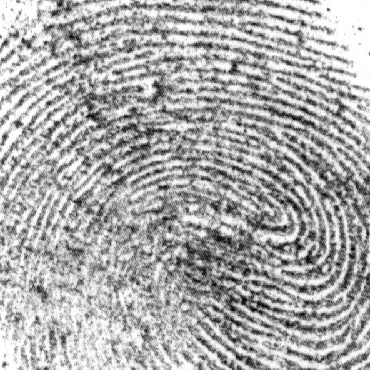} &  
             \includegraphics[height=2.25cm]{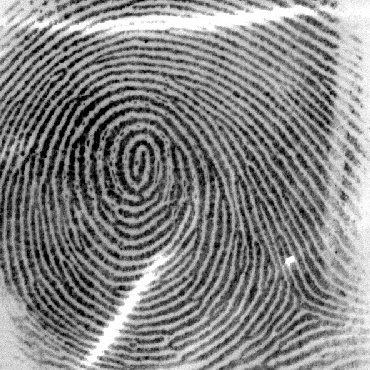}  &
             \includegraphics[height=2.25cm]{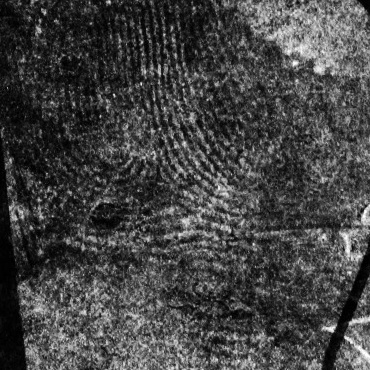} &
            \includegraphics[height=2.25cm]{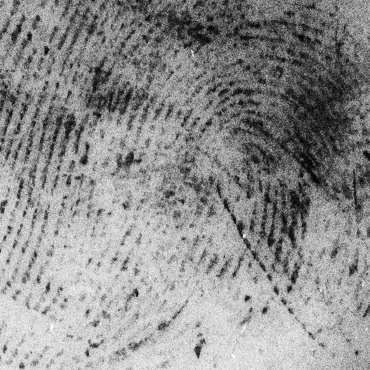}&
            \includegraphics[height=2.25cm]{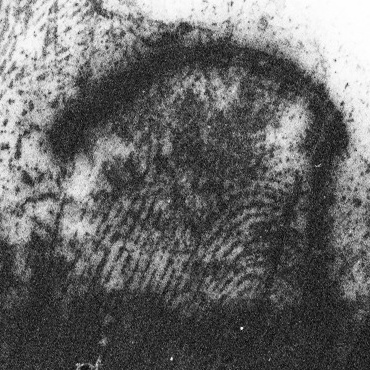} &
            \includegraphics[height=2.25cm]{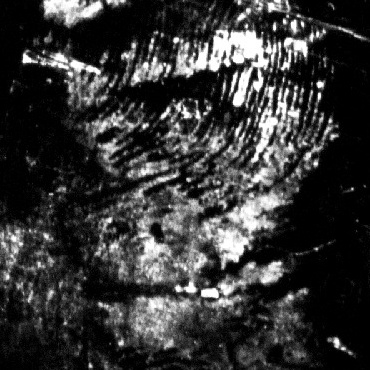}
             \\

             \begin{turn}{90}\footnotesize \tab[0.15cm]Synthetic latents\end{turn}
             \includegraphics[height=2.25cm]{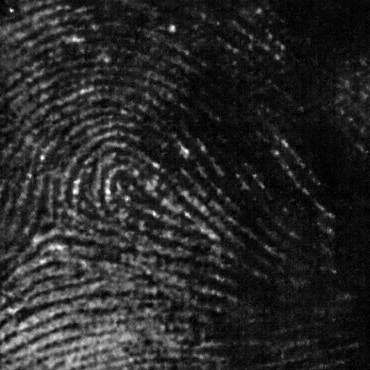} & 
             \includegraphics[height=2.25cm]{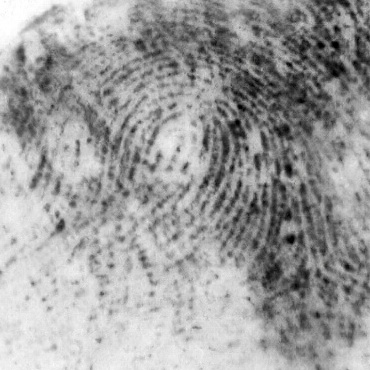} &
             
             \includegraphics[height=2.25cm]{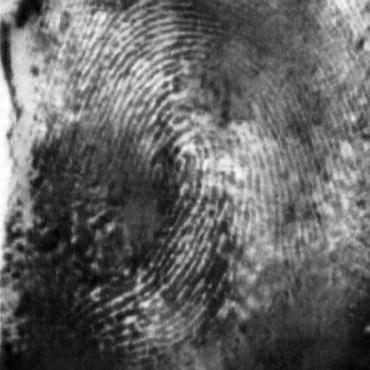} &
            \includegraphics[height=2.25cm]{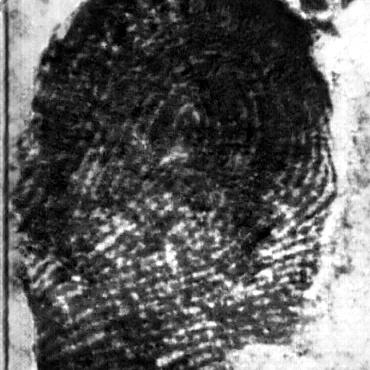} &

            \includegraphics[height=2.25cm]{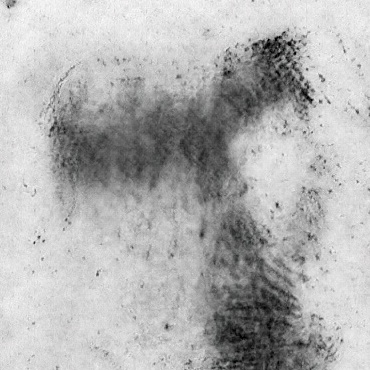} &
            \includegraphics[height=2.25cm]{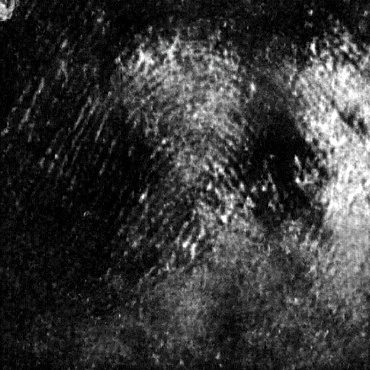} \\
            
		\end{tabular}

            \tab[0.3cm] Good \tab[4.6cm] Bad \tab[4.8cm]  Ugly

		\caption{Visual comparison between Good, Bad and Ugly latents from NIST SD27, MSP database and our synthetic latent prints (SLP). NIST SD27 database comes with a subjective categorization into three classes based on image quality (Good, Bad and Ugly). We use the same categorization of MSP latents and our synthetic fingerprints. There is a distinct difference in the recognition accuracies of these three classes of latent prints as shown in Figure \ref{roc_models}. }
		\label{visualcomparison}
\end{figure*}

\subsubsection{Generating multiple latent print styles}
\label{finetuneCycleGAN}

We use the CycleGAN model developed in the first stage to generate multiple latent print styles and finetune the $k$ clusters based on visual similarity. Each of the $k$ CycleGAN models generates a different variation of the characteristics present in each latent group. Therefore, if the number of latent images per cluster is small, CycleGAN may not converge well. Section \ref{SLF} details data augmentation operations that help the training converge better.

With $k$ finetuned CycleGAN models. Greater control can be exerted over the distribution of synthetic latent fingerprint styles. Our approach uses identities (represented as rolled fingerprints) obtained from the NIST SD4 database for validation in our experiments. However, any other rolled fingerprint database can be used, for example, even the synthetic rolled images from PrintsGAN  \cite{engelsma2022printsgan} for generating new synthetic identities. PrintsGAN is a neural method capable of generating unique identities by simulating parameters of distortions, textures, and finger pressure on a surface, among others. In combination, PrintsGAN and the $k$ CycleGAN texture models can generate a wide range of latent-rolled mated pairs of fingerprints with ``virtual" identities.  

\section{Experimental Results}

\subsection{Synthetic latent fingerprint generation}
\label{SLF}

During the first stage of our algorithm to create the latent fingerprint set, we train a CycleGAN model that receives as input 2,000 rolled fingerprints from the first acquisition of the NIST SD4 \cite{watson1992nist}. For our target texture transfer, we use 2,074 latent fingerprints from the MSP latent database. At the end of this first stage, we created a model capable of converting rolled images into latent images. 

Although this model trained in the first stage is already capable of generating latent fingerprints, we aim to generate greater diversity in the visual characteristics of the latent images, providing different matching performance behaviors. In the second stage, we choose 3 clusters ($k=3$) since the NIST SD27 latents have been partitioned into three categories, namely good, bad, and ugly, with three distinct levels of recognition 
difficulty and texture.

Given the three clusters in the first stage, we execute the second stage of our method, finetuning the coarse model with the styles for each cluster. We applied data augmentation operations on the fly during training by doing a random vertical and horizontal translation of the images (maximum of 100 pixels) and a random rotation with a maximum of 15 degrees. Finally, after training, we generated three CycleGAN latent print models with different characteristics, including distortion, texture and friction ridge area, a factor that provides more control in generating latent images. To define which models generate Good, Bad, or Ugly latents, we performed a recognition performance experiment presented in Section \ref{Similarity_experiment}. 

To create our final set of synthetic latent images, we used rolled images from the second acquisition of the NIST SD4 \cite{watson1992nist} and used the three CycleGAN models to generate 2,000 new latent images. These images comprise our database of 2,000 synthetic latents and their 2,000 rolled mated images from the first acquisition of the NIST SD4. Figure \ref{visualcomparison} shows a visual comparison between the good, bad, and ugly groups from the MSP latent database, NIST SD27, and our synthetic latent images.

\subsection{Similarity of real and synthetic latent embeddings}
\label{Similarity_experiment}

\begin{figure}[h]
\centering
            \includegraphics[height=6.3cm]{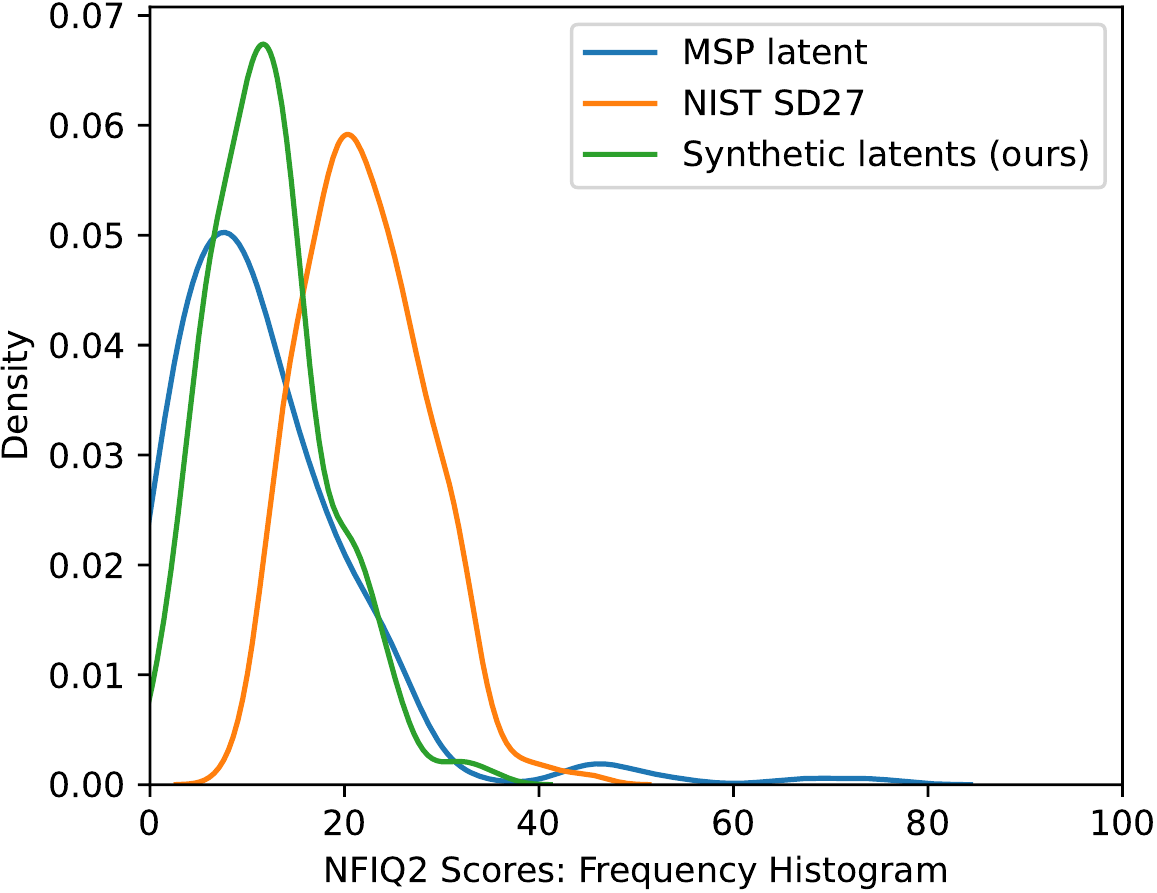}
            \caption{Histogram of NFIQ 2 scores for MSP latents, NIST SD27 and our synthetic latent prints. Note that the quality histogram of synthetic latents generated by our method overlaps the quality histograms of crime scene latents from NIST SD27 and MSP latents.}
            \label{nfiq2}
\end{figure}

To validate the similarities between our synthetic latents and real latent fingerprints in MSP latent and NIST SD27 databases, we first used NFIQ 2 \cite{bausinger2011fingerprint} to compare the quality of latents. NFIQ 2 provides a score in the range [0, 100], where higher the score, better the image quality. Figure \ref{nfiq2} contains the frequency histogram of the scores obtained from the MSP latents, NIST SD27, and our synthetic latent prints (SLP), showing that the quality of NIST SD27 latent images is, overall, better than MSP latents and our synthetic latents. The quality of our synthetic latents is similar to the MSP latent images, which is reasonable since we use the MSP latent in our training of CycleGAN.

\begin{figure}[h]
\centering
            \includegraphics[height=6.6cm]{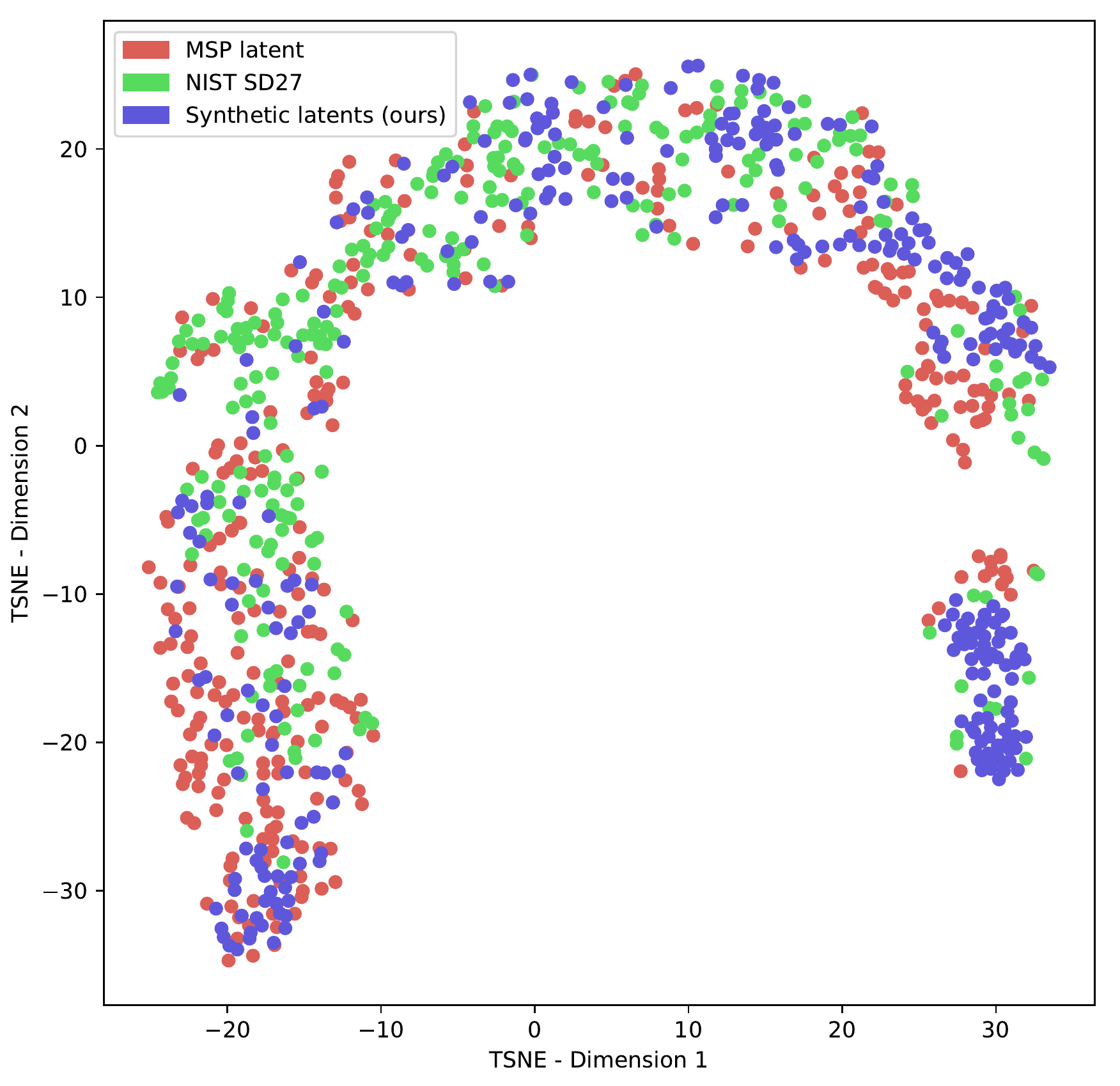}
            \caption{2D t-SNE plot of 192D embeddings showing the overlap of NIST SD27, MSP and synthetic latents images.}
            \label{TSNE}
\end{figure}

To further illustrate the similarity of synthetic latents to MSP latents and NIST SD27 latents, we show a t-Distributed Stochastic Neighbor Embedding (t-SNE) plot in Figure \ref{TSNE}. We see that the placement of 2D embeddings of latent fingerprints of real databases our and synthetic latents is similar. These results show that we are able to realistically generate latent fingerprints that replicate visual features of real latent images.

\begin{figure}[h]
\centering
            \includegraphics[height=6cm]{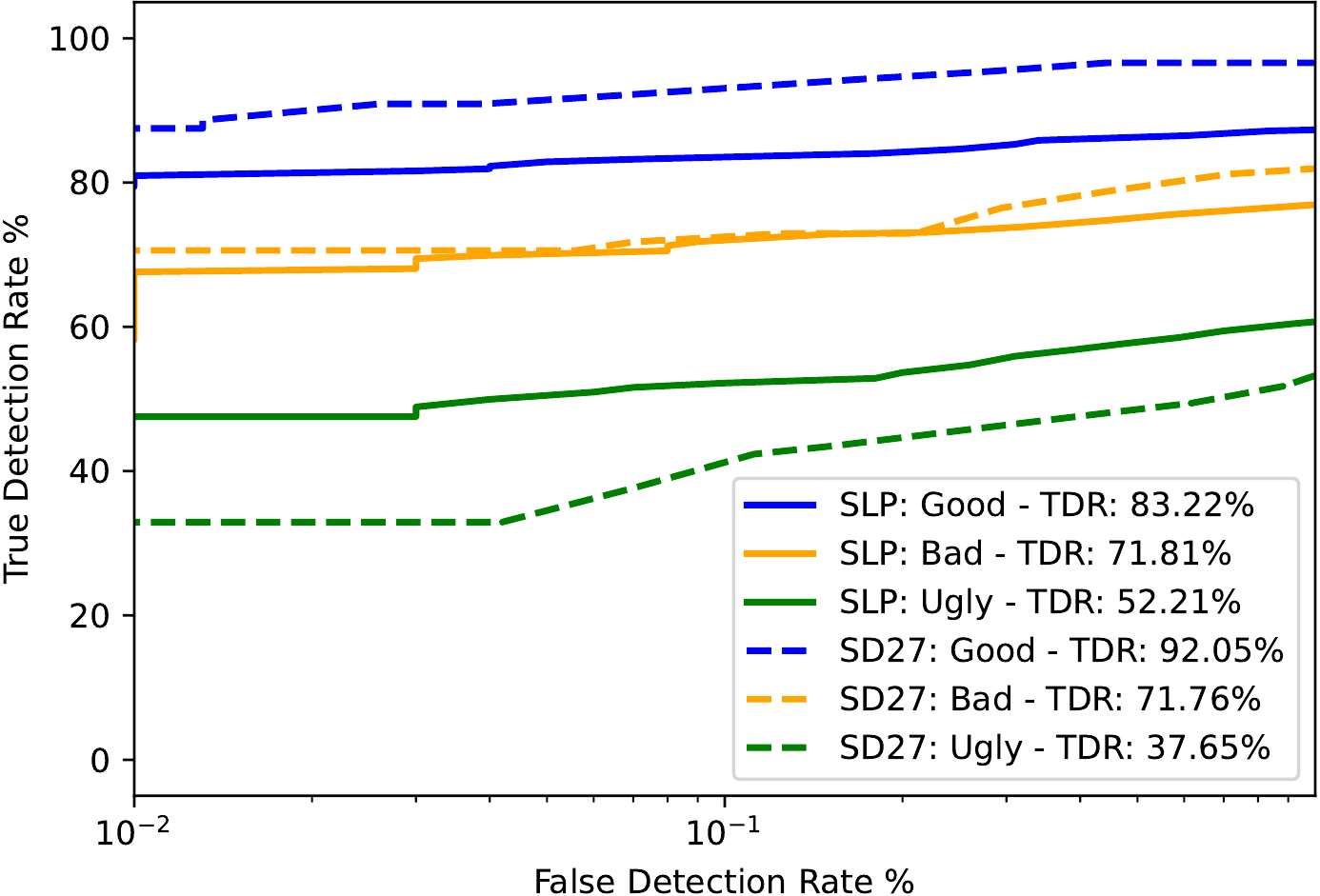}
            \caption{ROC curves for synthetic latents  prints (SLP) generated from NIST SD4 to their rolled mates in NIST SD4 and NIST SD27 latents to NIST SD27 rolled comparisons. Good, Bad and Ugly accuracy plots for the two comparisons appear to grouped together  in these plots. TDR (\%) reported @ FAR = 0.1\%.}
            \label{roc_models}
\end{figure}

Finally, for each of the three CycleGAN models, one each for Good, Bad and Ugly latents, we matched the synthetic latent images with their mated rolled images from the NIST SD4 database using Verifinger V12.3 SDK. Fig \ref{roc_models}. shows the ROC plots for the Good, Bad and Ugly groups of synthetic latents. For comparison, similar plots are shown for the NIST SD27 database. This comparison shows that the recognition difficulty of Good, Bad and Ugly synthetic prints is similar to the difficulty levels of the corresponding three types of latents in NIST SD27 latents.

\subsection{Minutiae analysis}

\begin{figure}[h]
	\centering
		\begin{tabular}{cc}
		
            \includegraphics[height=4.5cm]{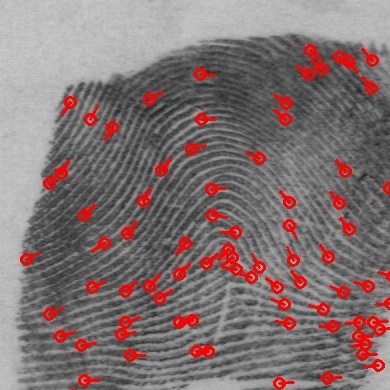} &  
            \includegraphics[height=4.55cm]{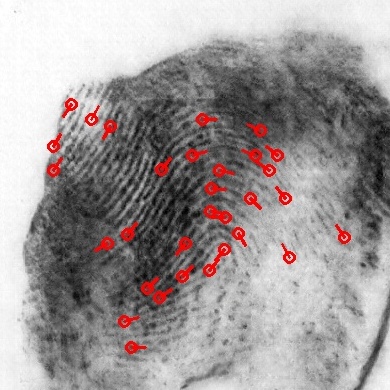} \\ 
            \footnotesize	 (a) NIST SD4 rolled (88) & \footnotesize	(b) Good latent from (a) (31)\\ 
            \includegraphics[height=4.5cm]{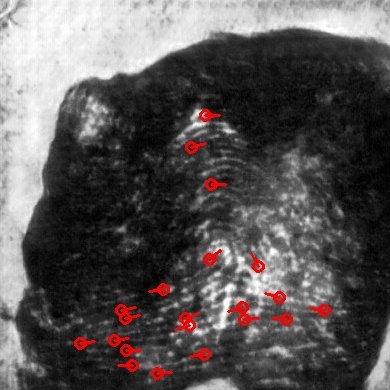} &
            \includegraphics[height=4.5cm]{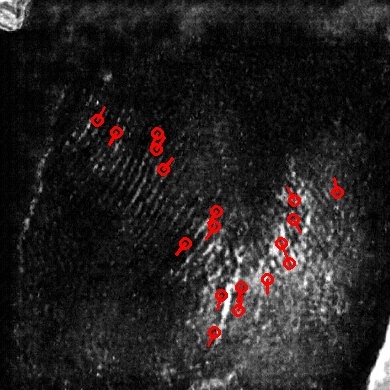} \\

            \footnotesize	(c) Bad latent from (a) (21)& \footnotesize (d) Ugly latent from (a) 
(19) 
            \end{tabular}
\caption{Good, Bad and Ugly latents synthesized from a rolled image in (a). No. of minutiae in the images (a)-(d) by Verifinger V12.3 SDK are shown in parentheses. Generally, lower the minutiae count, lower the image quality. }
            \label{min_img}
\end{figure}

To further analyze the visual differences between Good, Bad, and Ugly latent images generated by CycleGAN models, in Figure \ref{min_img}, we show Good, Bad and Ugly latent images generated from a single NIST SD4 identity, overlaid with the minutiae extracted by Verifinger V12.3. Note that the number of minutiae decreases as we observe images generated by Good, Bad and Ugly models. To make a quantitative comparison of minutiae count between real and synthetic latent prints, we generated 258 latent prints, the same as in NIST SD27. The mean minutiae count for the NIST SD27 are: 68, 45, and 35 for Good, Bad and Ugly latents. The corresponding minutiae count mean for our synthetic prints are: 55, 39, and 35, respectively.

\subsection{Data augmentation to increase latent recognition accuracy}

This experiment aims to show that it is possible to increase the matching accuracy of the rolled-to-rolled matcher DeepPrint \cite{engelsma2019learning} by using the synthetic latent prints (SLP) as data augmentation for retraining DeepPrint. In performing this experiment, we seek to ascertain the relative performance increase within the scope of DeepPrint and not relative to other matchers, thus validating the usefulness of synthetic latent images to improve the recognition performance of any fingerprint matcher for latent recognition. 

As a test set for our experimentation, we used NIST SD27 \cite{garris2000nist}, NIST SD302 (N2N) \cite{fiumara2019nist} and IIITD-SLF \cite{yusof2012multi}, performing the matching between rolled/slap images and latent images from each of these databases. Furthermore, as our method allows the controlled synthesis of Good, Bad and Ugly latents, we only utilize Bad and Ugly latents for data augmentation as these provide higher matching challenge. 

We designed the experiment as follows: (i) use the original DeepPrint model \cite{engelsma2019learning} as a reference model. (ii) finetune DeepPrint with MSP latent database images, finding an optimal value for the localization network hyper-parameter $LN$ in DeepPrint's architecture in the context of latent prints. Using the same $LN$ value as in (ii) allows direct comparison and finetuning DeepPrint with our synthetic SLP latent prints. (iii) generate a variation in the model that allows us to perform score level fusion, and perform finetuning using the same images as (ii), but with a different $LN$ value that still provides satisfactory recognition results. The models used in our experiments are summarized in Table \ref{models_training_localization}.

\begin{table}[H]
\centering
\caption{Models, training data and hyper-parameters.}
\label{models_training_localization}
\resizebox{0.7\columnwidth}{!}{%
\begin{tabular}{@{}ccc@{}}
\toprule
\textbf{Model}    & \textbf{Training data}                                                                                             & \textbf{\begin{tabular}[c]{@{}c@{}}Localization\\ network \\ hyper-parameter\end{tabular}} \\ \midrule
DeepPrint \cite{engelsma2019learning} & \begin{tabular}[c]{@{}c@{}}455K MSP rolled prints\end{tabular}                           & 0.035                                                                                      \\
$DeepPrint_1$      & \begin{tabular}[c]{@{}c@{}} 518 MSP aligned pairs of rolled and latents\end{tabular}            & 0.018                                                                                      \\
$DeepPrint_2$      & \begin{tabular}[c]{@{}c@{}}2K SD4 (rolled, synthetic latents) pairs (bad, ugly)\end{tabular} & 0.018                                                                                      \\
$DeepPrint_3$      & \begin{tabular}[c]{@{}c@{}}2K SD4 (rolled, synthetic latents) pairs (bad, ugly)\end{tabular} & 0.007                                                                                      \\ \bottomrule
\end{tabular}
}\\
\scriptsize	Note that $DeepPrint_2$ and $DeepPrint_3$ models do not use any real latent print for retraining.
\end{table}

\begin{figure}[H]
\centering
            \includegraphics[height=5cm]{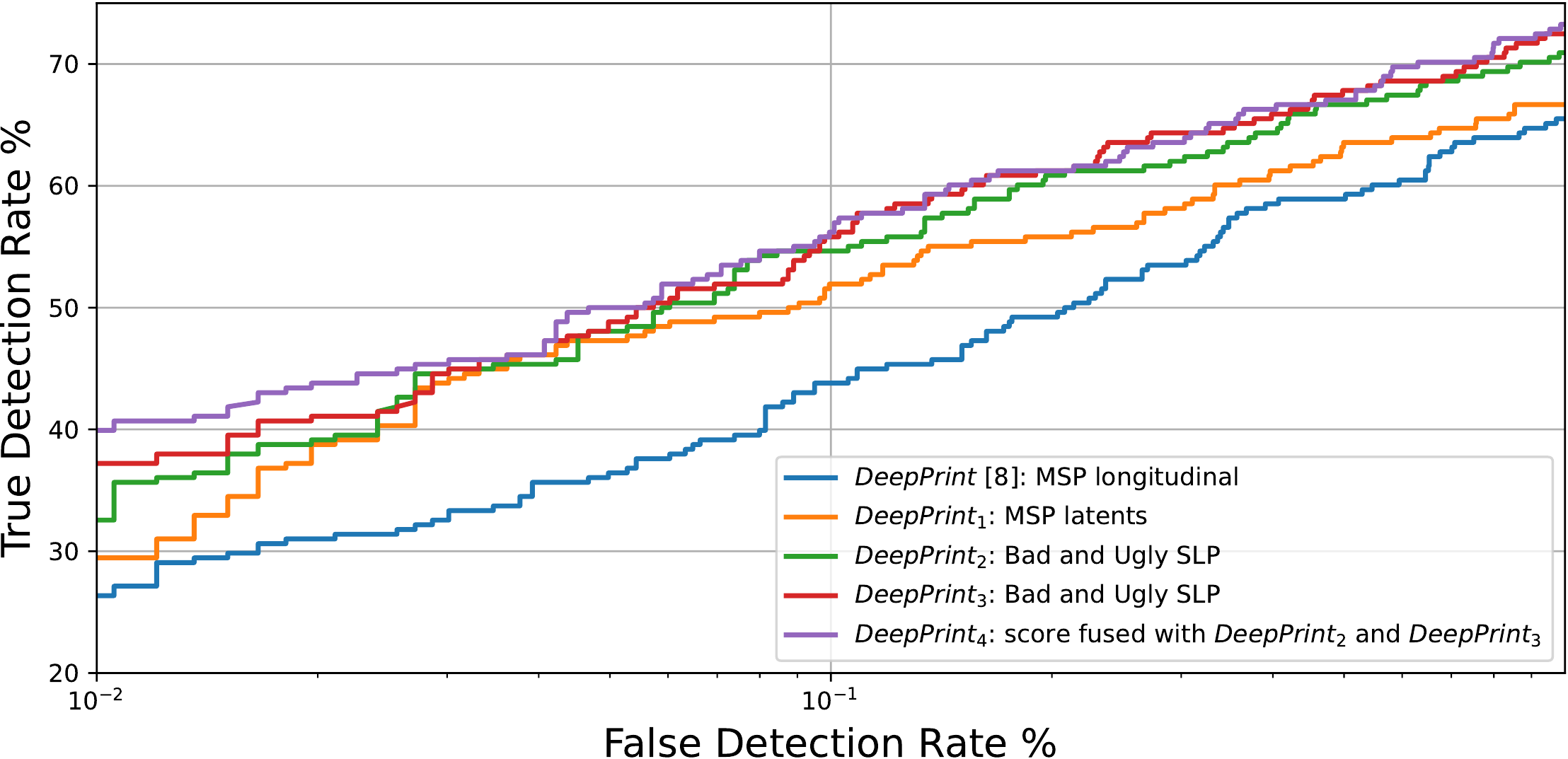}
            \caption{ROC curves of the DeepPrint \cite{engelsma2019learning}, $DeepPrint_1$, $DeepPrint_2$, $DeepPrint_3$ and $DeepPrint_4$ models evaluated on NIST SD27 \cite{garris2000nist}.}
            \label{deepprint_matching_1}
\end{figure}

In addition to the models in Table \ref{models_training_localization}, we applied Min-Max normalization to the scores generated by the $DeepPrint_2$ and $DeepPrint_3$ models and performed simple score level fusion \cite{Ross2009} as $ScoreFusion = \frac{(ns_2+ ns_3)}{2}$, where $ns_2$ and $ns_3$ are the normalized scores from $DeepPrint_2$ and $DeepPrint_3$. The ``fused" model is referred to as $DeepPrint_4$. Figure \ref{deepprint_matching_1} shows the ROC curve of the DeepPrint \cite{engelsma2019learning}, $DeepPrint_1$, $DeepPrint_2$, $DeepPrint_3$ and $DeepPrint_4$ models evaluated on NIST SD27 \cite{garris2000nist}. 

In all cases, our models trained with the synthetic images performed better than the original DeepPrint model and the finetuned models trained with the MSP latents on NIST SD27. We arrange all the experimental analysis results in Table \ref{augtable}. Only with the latent MSP latents was it possible to increase the system's performance accuracy. Still, it was necessary to perform alignment between the rolled and latent mates, which is also a challenging task. This alignment is needed because the Localization network in DeepPrint cannot perform such a precise alignment with latent images. However, the alignment problem does not exist with synthetic images since we can control the latent synthesis directly from a rolled print. These results corroborate that Verifinger V12.3 is one of the best-performing matchers, consistent with the results reported in FVC-onGoing \cite{dorizzi2009fingerprint} and NIST FpVTE \cite{wilson2004fingerprint}.

\begin{table}[H]
\centering
\caption{True detection rate (TDR (\%)) @ FAR = 0.01\%}
\label{augtable}
\resizebox{0.6\textwidth}{!}{%

\begin{tabular}{@{}cccc@{}}
\toprule
\textbf{Model}                       & \textbf{\begin{tabular}[c]{@{}c@{}}NIST SD27\end{tabular}} & \textbf{\begin{tabular}[c]{@{}c@{}}NIST SD302 (N2N)*\end{tabular}} & \textbf{\begin{tabular}[c]{@{}c@{}}IIITD-SLF\end{tabular}} \\ \midrule
DeepPrint \cite{engelsma2019learning}                   & 26.35                                                                      & 10.81 &  10.0                                                                        \\
$DeepPrint_1$                         & 29.45                                                                      & 13.15 &  14.16                                                                       \\
$DeepPrint_2$                         & 32.55                                                                      & 13.97 &  15.0                                                                        \\
$DeepPrint_3$                         & 37.20                                                                      & 14.07 &  17.5                                                                        \\
$DeepPrint_4$     & \textbf{39.92}                                                             & \textbf{14.31} &  \textbf{20.83}                                                                        \\
Verifinger V12.3 & \textbf{55.81}                                                             & \textbf{16.22} &  \textbf{29.70}                                                                        \\ \bottomrule
\end{tabular}
}
\\\scriptsize *N2N latent and rolled mates present in the finger position annotation in the SD302h subset. We reduced the resolution of N2N latent images to 500dpi and applied the Clahe filter \cite{pizer1987adaptive} to highlight the fingerprint ridges. 
\end{table}

\subsubsection{Identification (1:N Comparison)}

\begin{figure}[H]
\centering
            \includegraphics[height=5.1cm]{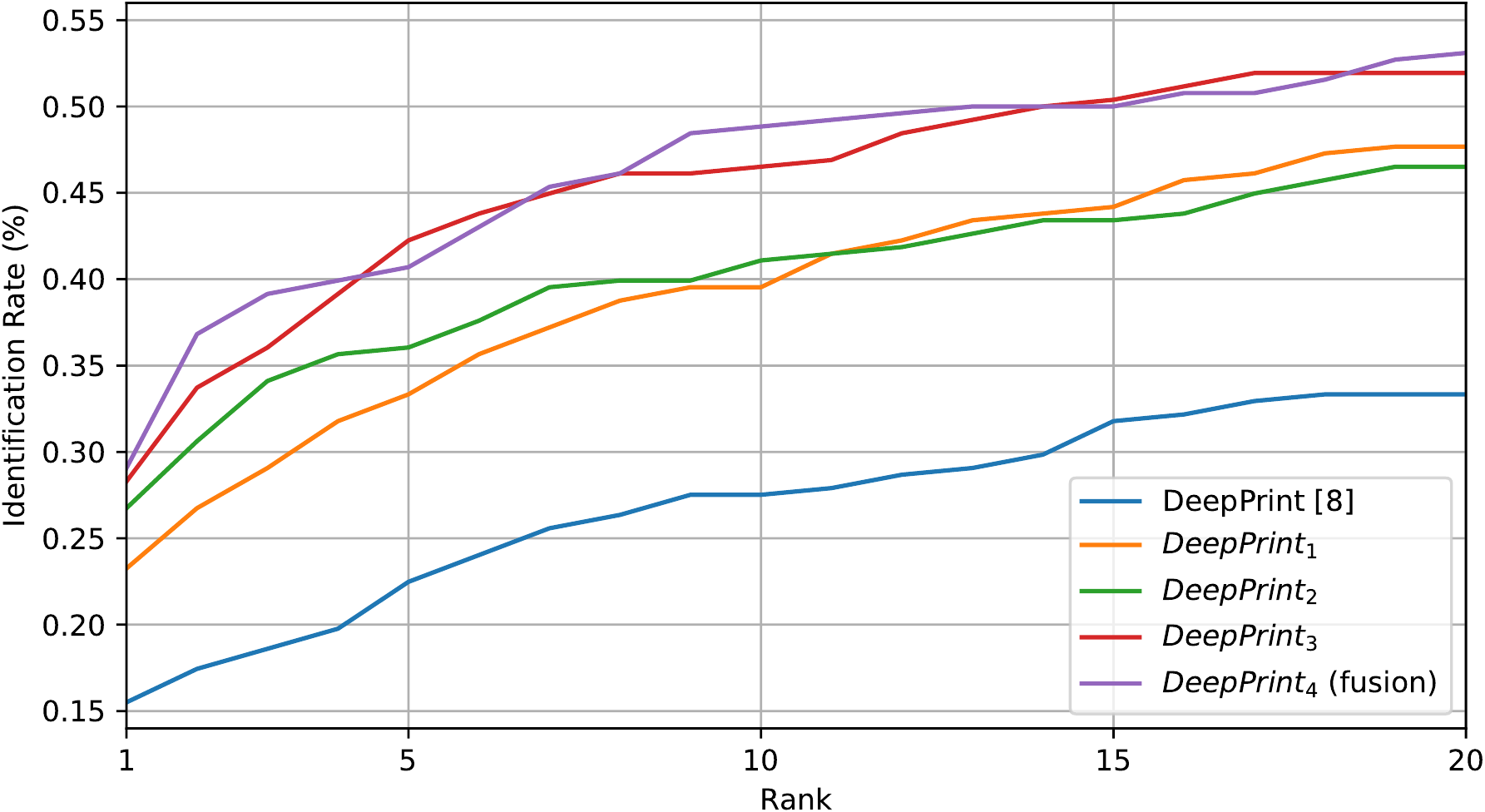}
            \caption{CMC curves of various DeepPrint and its variants evaluated on NIST SD27 latent database.}
            \label{cmc}
\end{figure}

Our goal of this experiment is to perform a closed-set analysis of identification accuracy by matching latent prints in NIST SD27, NIST 302 (N2N) and IIITD-SLF against a galley augmented by  62,871 rolled fingerprints from the NIST SD300a (8,871k) \cite{fiumara2018nist} and NIST SD14 \cite{watson2001nist} (54K) databases. The results of this analysis are shown in Table \ref{table_1_n} and the Cumulative Matching Characteristics (CMC) of the NIST SD27 is presented in Figure \ref{cmc}. We used the same protocol as NIST ELFT-EFS \cite{flanagan2010nist} to evaluate the performance of our models. 

\begin{table}[H]
\centering
\caption{Rank-1 accuracies (\%) of different DeepPrint models against a background set of 62,871 rolled fingerprints (54K from NIST SD14 and 8,871K from NIST SD300a), in addition to the true rolled mates of the latents prints of each database.}
\label{table_1_n}
\resizebox{0.6\textwidth}{!}{%

\begin{tabular}{cccc}
\hline
\textbf{Model}    & \textbf{NIST SD27} & \textbf{NIST SD302 (N2N)*} & \textbf{IIITD-SLF} \\ \hline
DeepPrint \cite{engelsma2019learning} & 15.50              & 4.18                      & 9.8              \\
$DeepPrint_1$      & 23.64              & 5.98                      & 14.13             \\
$DeepPrint_2$      & 26.74              & 6.56                      & 14.7              \\
$DeepPrint_3$      & 28.29              & 7.24                      & 17.12             \\
$DeepPrint_4$      & \textbf{29.07}     & \textbf{7.93}             & \textbf{18.21}     \\ 
Verifinger V12.3     & \textbf{45.45}     & \textbf{10.81}             & \textbf{22.71}     \\ 
\hline
\end{tabular}
}
\\ \scriptsize *We used all 9,990 latents in N2N. We reduced the resolution of N2N latent images to 500dpi and applied the Clahe filter \cite{pizer1987adaptive} to highlight the fingerprint ridges. 

\end{table}

\section{Conclusion}

We present a new approach capable of synthesizing latent fingerprints from any rolled fingerprint database. To accomplish this, we used a modified version of CycleGAN by adding a patch discriminator in addition to the global discriminator already present in the original CycleGAN. Our approach generates models capable of replicating Good, Bad and Ugly latent images present in the NIST SD27 latent database, thus allowing the generation of style-controlled latent images.  

Additionally, by comparing NFIQ 2 quality values and t-SNE plots in two dimensions and the recognition performance of the latent to their mated rolled images we can claim that the synthetic latent images are realistic and similar to the real latent images. 

Finally, we performed a finetuning on the original DeepPrint \cite{engelsma2019learning} model using our SLP latent images generated with our modified CycleGAN model.  We achieved a performance boost, validating this on three latent databases. In future work, we plan to refine the synthesis of latent prints by adding more complexity to background and texture variations. Furthermore, CycleGAN is capable of reverting a latent into a rolled one. With this, we plan to improve the quality of latent prints, thus improving latent recognition in deep networks.

\bibliographystyle{unsrtnat}
\bibliography{references}

\end{document}